# Finding Bottlenecks: Predicting Student Attrition with Unsupervised Classifier

Seyed Sajjadi, Bruce Shapiro, Christopher McKinlay, Allen Sarkisyan, Carol Shubin, Efunwande Osoba
California State University, Northridge
seyed.sajjadi.947@my.csun.edu, bruce.e.shapiro@gmail.com, chris.mckinlay@gmail.com, programminglinguist@gmail.com,
carol.shubin@csun.edu, eosoba@gmail.com

*Abstract*—**With pressure to increase graduation rates and reduce time to degree in higher education, it is important to identify at-risk students early. Automated early warning systems are therefore highly desirable. In this paper, we use unsupervised clustering techniques to predict the graduation status of declared majors in five departments at California State University Northridge (CSUN), based on a minimal number of lower division courses in each major. In addition, we use the detected clusters to identify hidden bottleneck courses.**

## I. Introduction

Policy makers, the public, university administrators, students and their families are concerned about low graduation rates and lengthy times to degree in higher education. The median time to graduation is six years at CSUN (1). The four-year and the six-year graduation rates are 13% and 50%, respectively (2). With an enrollment of over 6000 undergraduate students, CoBaE is one of largest business schools in the nation. CoBaE confers the second most undergraduate degrees at CSUN (behind the College of Social and Behavioral Science), and it has three of the top ten most popular majors (Management, Finance, and Marketing) at CSUN. We focused our analysis on three departments within the CoBaE because of the commonality of core curriculum. We also studied two departments from the College of Engineering because of pre-requisite curriculum.

We trained K-means classifiers on grade data from undergraduate majors in Business Law, Management, Marketing, Civil and Electrical Engineering. We found strongly predictive clusters in each of the five departments. Cluster separation was driven disproportionately by a small number of courses which we consider the bottlenecks to graduation. In fact, the first three classes of the graduation pathway give an effective early indication of student success or failure.

## II. Related Work

Educational data mining is an emerging discipline, concerned with developing methods for exploring the unique types of data that come from the educational sphere. The field encompasses various subdomains such as modeling student learning to better optimize performance, to detecting outliers, to developing automated tutoring systems that intelligently adapt lesson plans to the individual learning styles (11).

Luan (9) studied clustering aspects of data mining and offers comprehensive characteristic analysis of students and likelihood estimates for a variety of outcomes such as transferability, persistence, retention, and success in classes. Al-Radaideh et al. (3) applied classification techniques to determine the main attributes that may affect student performance. Tair and El-Halees (13), used K-means to predict graduate students' performance, and overcome the problem of low grades of graduate students. Ayesha, Mustafa, Sattar and Khan (4) have also used K-means clustering to predict student performance in a particular course. Romero, Ventura and Garca (12) described the full process of clustering, classification, visualization and statistics in the context of mining Moodle (e-learning) data. Our current work uses unsupervised clustering methods to address the issue of large scale student behavior, and attempts to identify student successes and failure through predictive clustering.

## III. Method

### A. Data Collection and Preprocessing

We obtained academic records containing grade information from declared majors in five departments in the College of Business and Economics and College of Engineering at CSUN. The majors we inspected were Management, Marketing and Business Law, Civil and Electrical Engineering. The data spans a ten-year period between 2004 and 2014 containing 9,088 student records in total and contains only the courses required for each major. The grade data for each course were encoded with the following normalized GPA scale prior to statistical analysis:

TABLE I. Grade Encoding Scheme

| A   | A-  | B+  | B   | B-  | C+  | C   | C-   | D+   | D    | D-   | F    |
|-----|-----|-----|-----|-----|-----|-----|------|------|------|------|------|
| 2.0 | 1.7 | 1.3 | 1.0 | 0.7 | 0.3 | 0.0 | -0.3 | -0.7 | -1.0 | -1.3 | -2.0 |

Not taking a required course for a specific major will prevent students from graduating. This has the same effect as failing the course; therefore, such a course was assigned a grade of 'F'. The datasets were separated by majors, with columns for graduation, number of semesters in the major, number of credits for the major, number of transfer credits earned, followed by the course names.

## B. *Cluster Analysis*

There are a few fundamental issues involved in cluster analysis, notably determining whether discrete clusters are present (8) and choosing the appropriate number of clusters (7) (6). We applied the K-means algorithm (10) to the grade data and used the Calinski-Harabasz (CH) index (5) to determine the optimal number of clusters on fivefold cross-validated datasets (Figure 1).

We then established the predictive power of the clusters by testing them on a classification task. We compared the cluster-based classifiers with logistic regression classifiers in predicting the graduation status of held-out samples for each department. A very common technique to measure the classifier performance is Receiver Operating Characteristic (ROC) curves which display the sensitivity of the model by plotting the true positive rate versus the false positive predictions and depicting their relative trade-offs. We then used ROC curves to evaluate and compare predictive performance of clustering and logistic regression methods for each department (Table 2). Finally, we performed the same steps to predict graduation status based on the first three courses in each major.

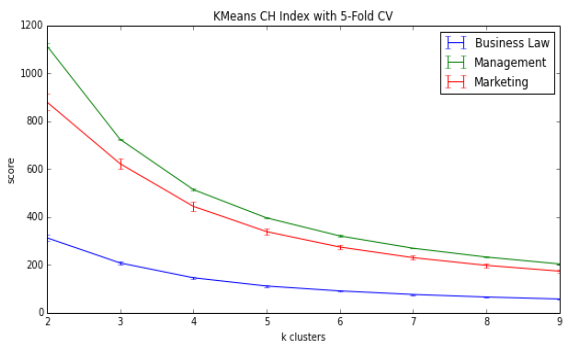

Fig. 1.   Calinski-Harabasz indices for all three datasets

TABLE II.   ROC P<small>LOTS FOR</small> B<small>OTH</small> C<small>LASSIFIERS ON THE</small> F<small>ULL</small> C<small>OURSE</small> S<small>ET AND THE</small> F<small>IRST</small> T<small>HREE</small> C<small>OURSES FOR</small> B<small>USINESS</small> M<small>AJORS AND</small> E<small>NGINEERING</small> M<small>AJORS</small>

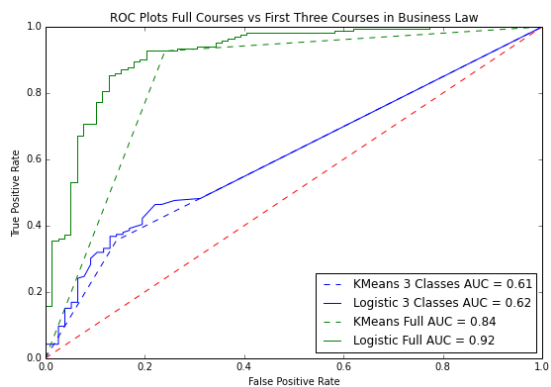

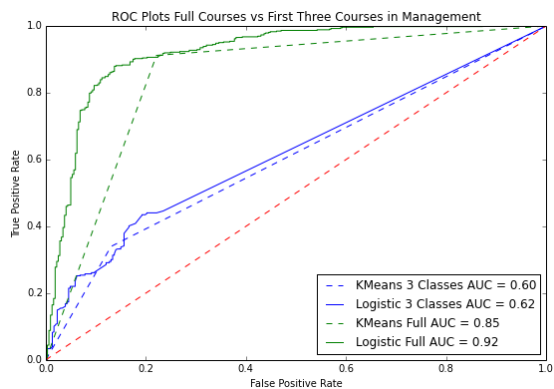

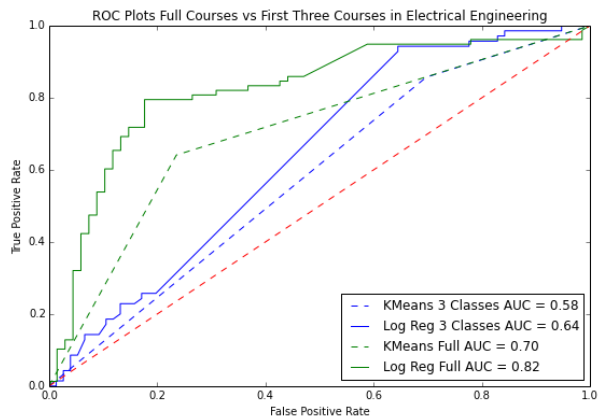

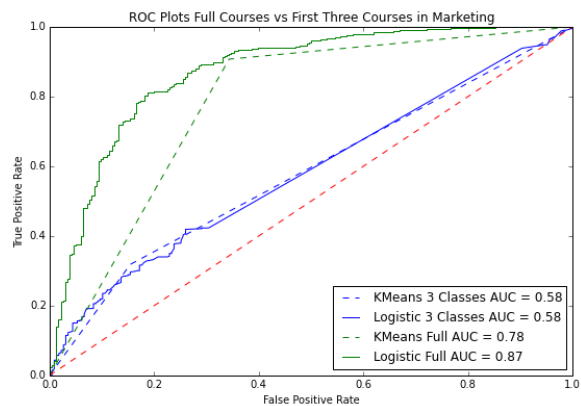

TABLE III.   A<small>CCURACY</small>, P<small>RECISION</small>, R<small>ECALL</small>, <small>AND</small> F1 S<small>CORES</small>

| | Model | Accuracy | Precision | Recall | F1 Score |
|---|---|---|---|---|---|
| Business Law | Logistic 3 courses | 0.62 | 0.62 | 1.0 | 0.77 |
| | Kmeans 3 Courses | 0.55 | 0.84 | 0.34 | 0.48 |
| | Logistic Full set | 0.86 | 0.87 | 0.92 | 0.90 |
| | K-means Full set | 0.86 | 0.87 | 0.92 | 0.89 |
| Electrical Engineering | Logistic 3 courses | 0.58 | 0.57 | 0.9 | 0.70 |
| | Kmeans 3 Courses | 0.54 | 0.55 | 0.82 | 0.66 |
| | Logistic Full set | 0.77 | 0.82 | 0.72 | 0.77 |
| | K-means Full set | 0.77 | 0.82 | 0.72 | 0.77 |
| Management | Logistic 3 courses | 0.61 | 0.61 | 1.0 | 0.76 |
| | Kmeans 3 Courses | 0.56 | 0.81 | 0.36 | 0.50 |
| | Logistic Full set | 0.85 | 0.88 | 0.88 | 0.88 |
| | K-means Full set | 0.83 | 0.84 | 0.90 | 0.87 |
| Civil Engineering | Logistic 3 courses | 0.59 | 0.58 | .88 | 0.70 |
| | Kmeans 3 Courses | 0.56 | 0.56 | 0.85 | 0.67 |
| | Logistic Full set | 0.84 | 0.87 | 0.83 | 0.85 |
| | K-means Full set | 0.76 | 0.84 | 0.70 | 0.76 |
| Marketing | Logistic 3 courses | 0.71 | 0.71 | 1.0 | 0.83 |
| | Kmeans 3 Courses | 0.55 | 0.67 | 0.71 | 0.69 |
| | Logistic Full set | 0.84 | 0.88 | 0.89 | 0.88 |
| | K-means Full set | 0.83 | 0.86 | 0.90 | 0.88 |

## IV. RESULTS

The optimal number of clusters was two in all departments, s determined by CH-index (Figure 1). The Management and Marketing departments showed better between-cluster separation than Economics and Business Law. We applied the same approach to the first three classes that students would normally take within their first year at school. (Table 3) shows the Accuracy, Precision, Recall and F1 Scores resulting from the two classifiers when trained on the full feature set (approximately 113 courses) and on the first three courses in each major.

We expected that a predictive model trained on the full feature set of course grades would be more effective than a model using cluster labels from unsupervised clustering. To test this hypothesis, we compared the performance of a logistic regression classifier trained on the full feature set to the performance of a classifier that used co-membership information from the clusters on a classification task: to predict whether the student had in fact graduated with that major. The cluster-based classifier estimated the probability that a student belonged to a particular category using the fraction of co-clustered samples that also belonged to the category of interest.

In each case we identified strongly predictive clusters. Though outperformed, the cluster- based classifiers compared surprisingly well with the logistic regression models (Table 2). Heuristically speaking, students in the same cluster tended to drop out at the same times after getting the same grades in the same courses.

## V. DISCUSSION

Cluster analysis can also help to identify common traits among students within each cluster. For each department the second cluster spends on average four semesters enrolled with that major declared (Table 4). However, the probability of these students graduating with the major is quite low (Table 3). Bottlenecks can be visually depicted as courses which are most well separated by predictive clusters. The average course grades seen in (Table 5) are further apart between clusters for upper division courses, as expected due to their requirement for graduation. Examination of the lower division course work can be seen as the beginning of the bottleneck for each major. Lower division coursework with relatively high separation in average grades between clusters are the best indicators for the separation between students that graduate and those which do not, and therefore act as bottlenecks in the major. These courses are excellent features to include into an early-warning system classifier as they are the most well separated vectors.

## VI. LIMITATIONS AND FUTURE WORK

Students may fail to graduate in CoBaE because they either change majors or discontinue their education at CSUN. Hierarchical clustering methods could provide more detailed information on student outcomes, such as predicting which department a student might change their major to. Collaborative filtering methods could also give departmental recommendations to students considering a change of major. These methods could be used to develop early warning and recommendation systems for automated advisement, which would be especially beneficial to over-taxed advisement systems at comprehensive state universities such as CSUN.

Another related issue is the incidence of major-switching is an important factor, since re-declaring a major is time-consuming and costly. Approximately 24% of CSUN students re-declare their major, and the plurality of these changes involve departments in the David Nazarian College of Business and Economics (CoBaE) (1).

Our results can be further refined by adding student metadata, for example: college year the major was declared, number of transfer credits, number of classes per term, financial aid, student demographics (such as age, gender, ethnicity, zip code) and various measures of student preparedness like SAT scores. With a more detailed feature space, our methods might be able to identify patterns and more well defined clusters.

## VII. CONCLUSION

We trained unsupervised classifiers on grade data from four undergraduate majors at CSUN. In each case we found strongly predictive clusters, and found that cluster separation was driven disproportionately by a small number of bottleneck courses. We also found that training classifiers on the first three classes on the graduation pathway was an effective early detection method. We argue that reforming, or at the very least investigating, these bottleneck courses are crucial to understanding student attrition

### ACKNOWLEDGMENTS

We gratefully acknowledge support from CSUN's Office of the Provost and Academic Affairs. We thank Provost Harry Hellenbrand and Vice-Provost Michael Neubauer for their institutional support, and Bettina Huber, CSUN Director of Institutional Research, for making the data available. We also thank Yauheniya (Gina) Lahoda and Dr. Bruce Shapiro for numerous conversations during the Spring 2015 Machine Learning seminar.

TABLE IV.  NUMBER OF SEMESTERS AND UNITS SPENT AT CSUN ACROSS THREE MAJORS OF BUSINESS LAW, MARKETING AND MANAGEMENT, ALSO SHOWING TRANSFER CREDITS FROM OTHER INSTITUTES WITH CLUSTER "1" REFERS TO THE STUDENTS WHO GRADUATED AND CLUSTER "2" TO THOSE WHO DID NOT

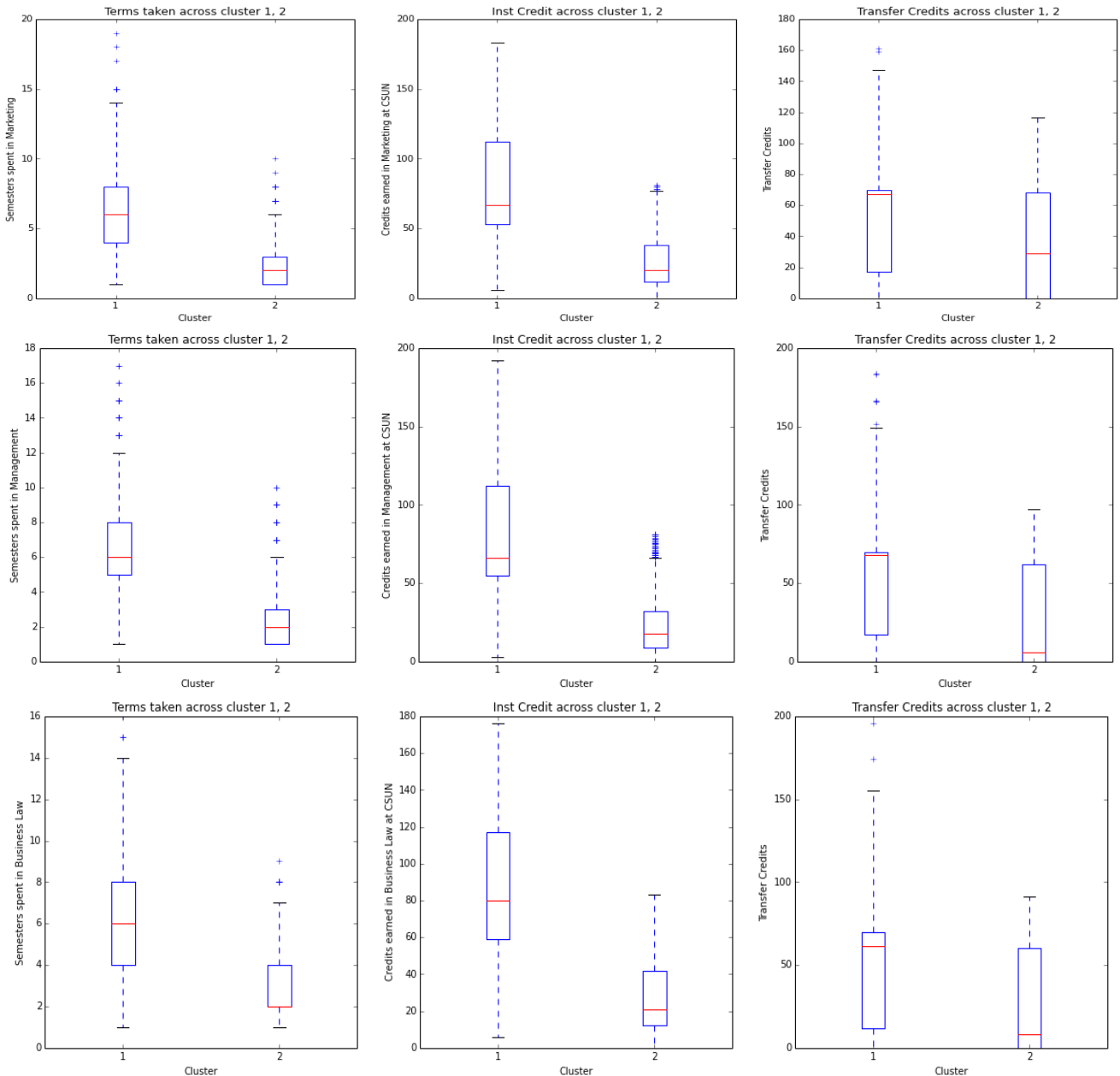

TABLE V. AVERAGES GRADES FOR EACH CLASS BY CLUSTER FOR BUSINESS AND ENGINEERING MAJORS

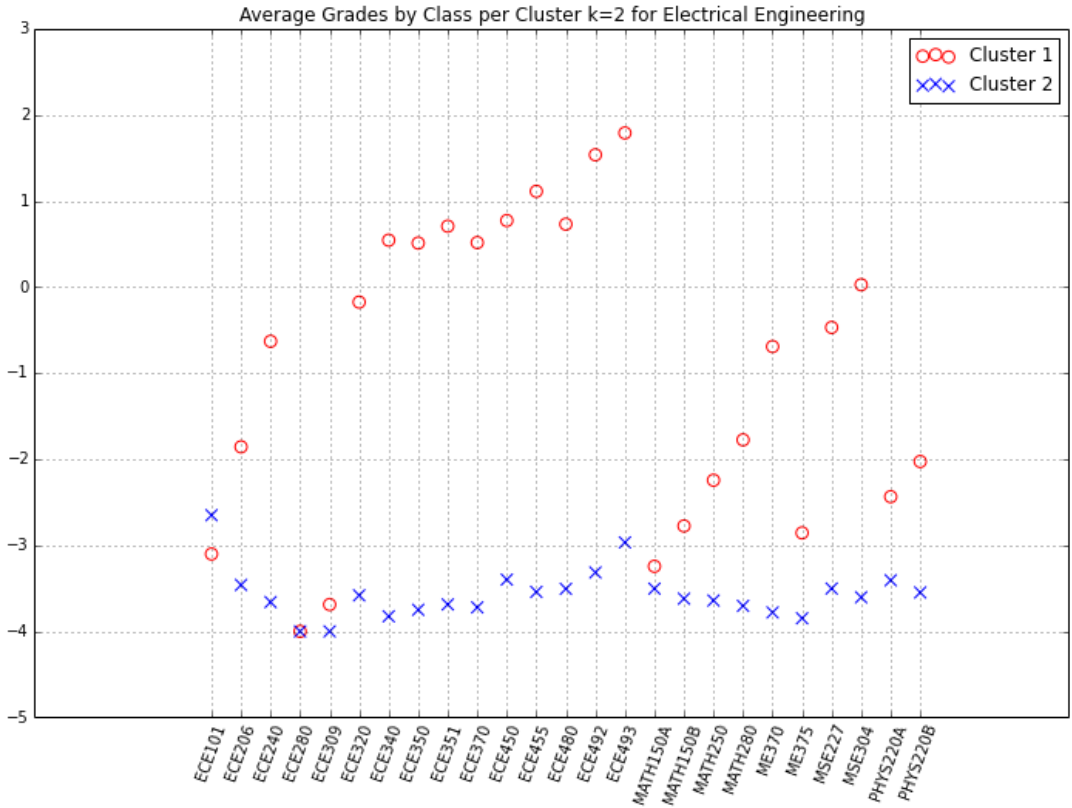

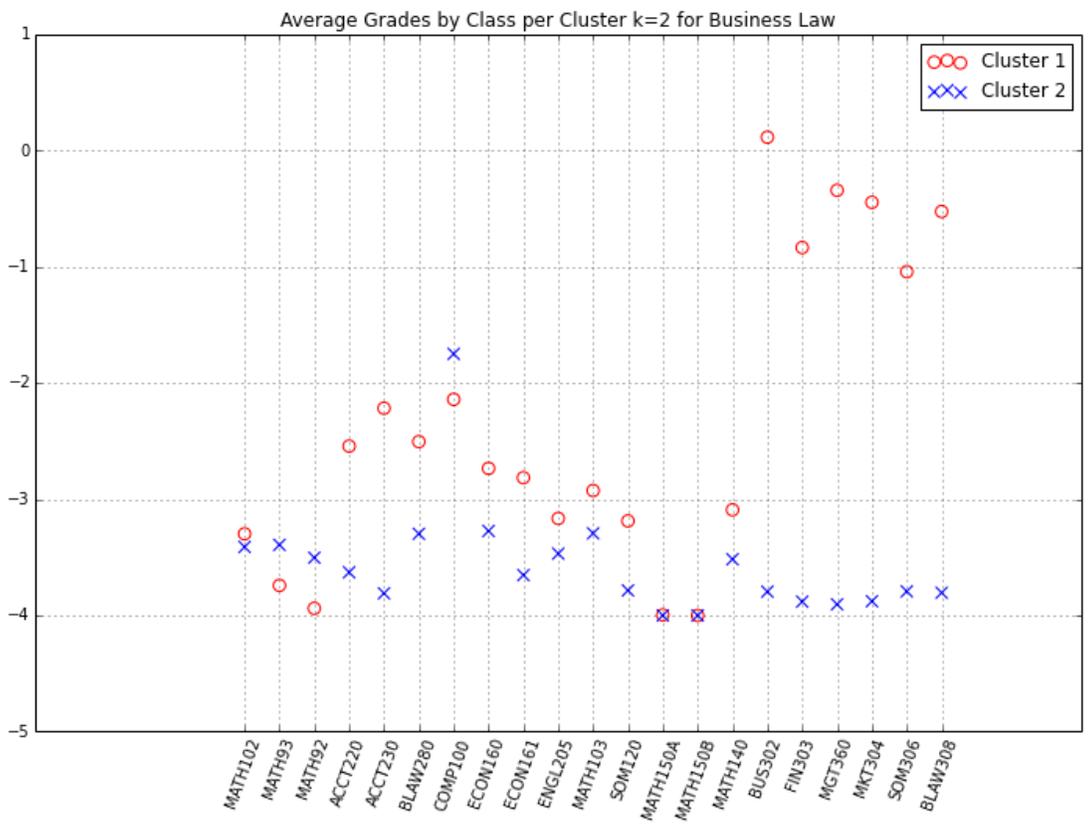
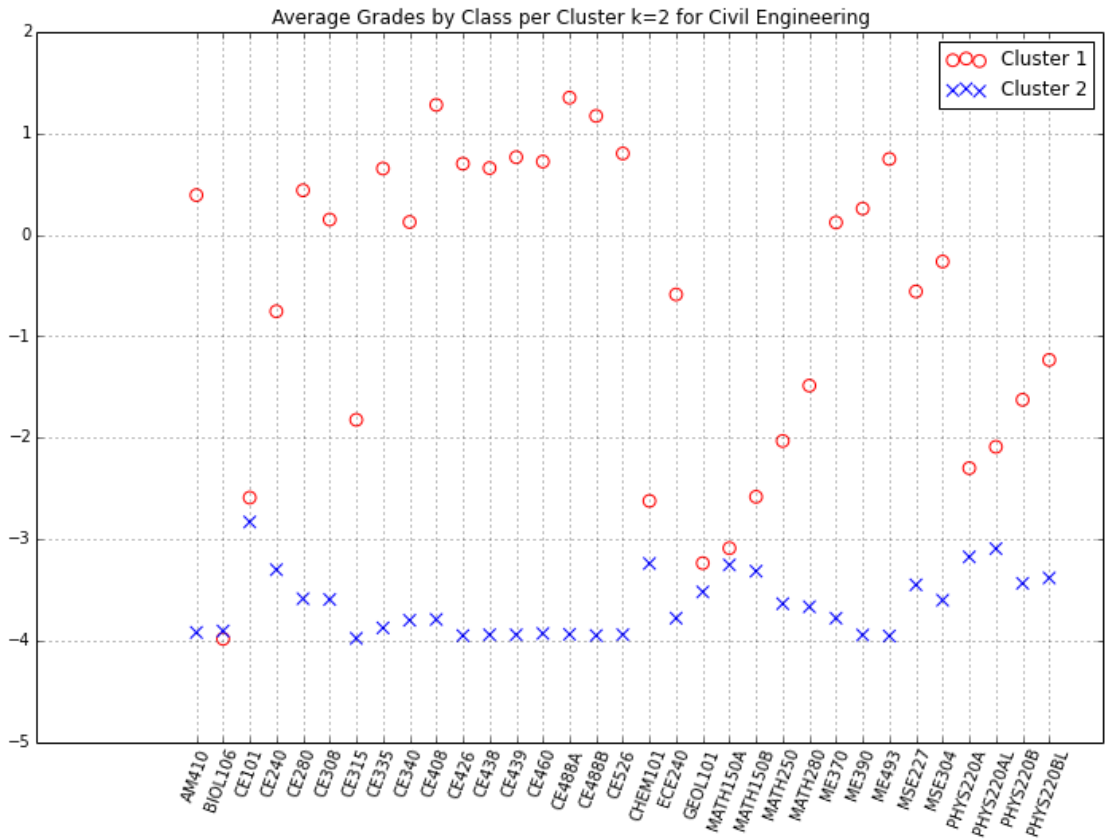

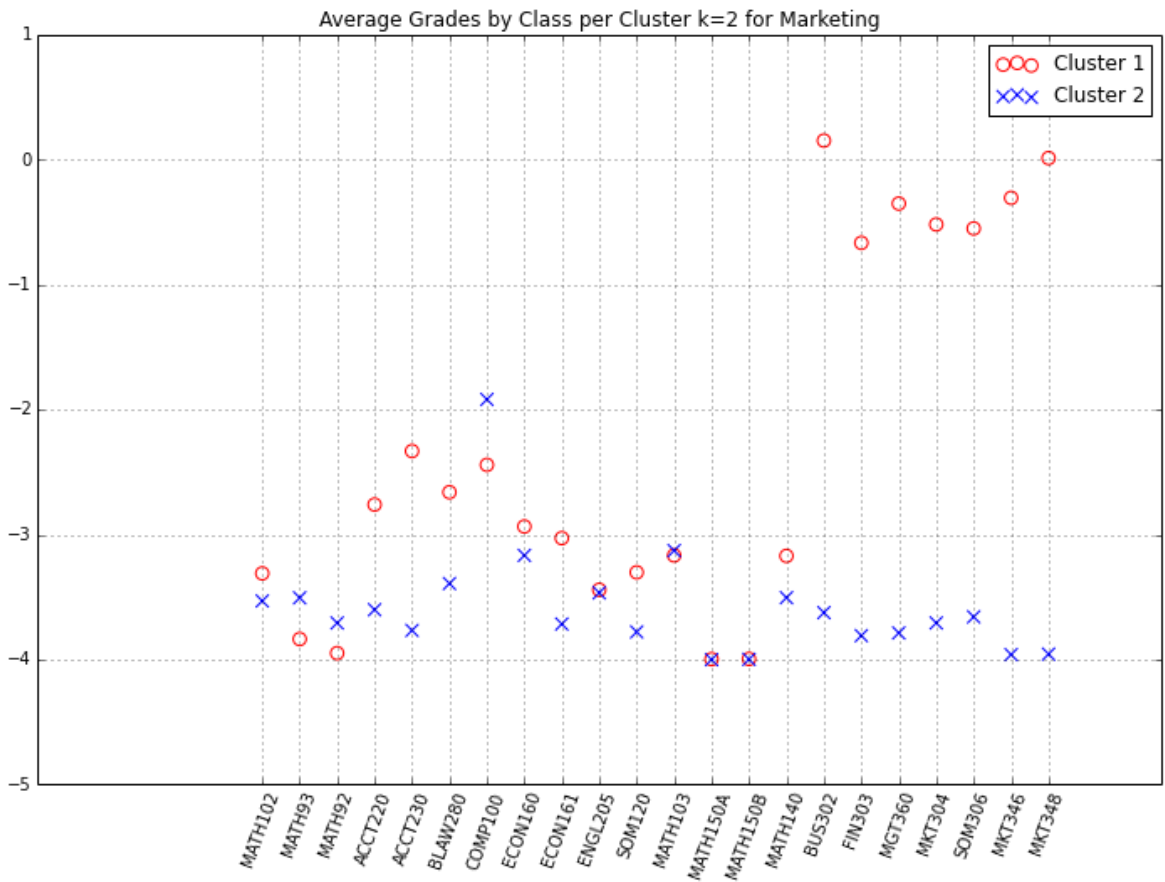
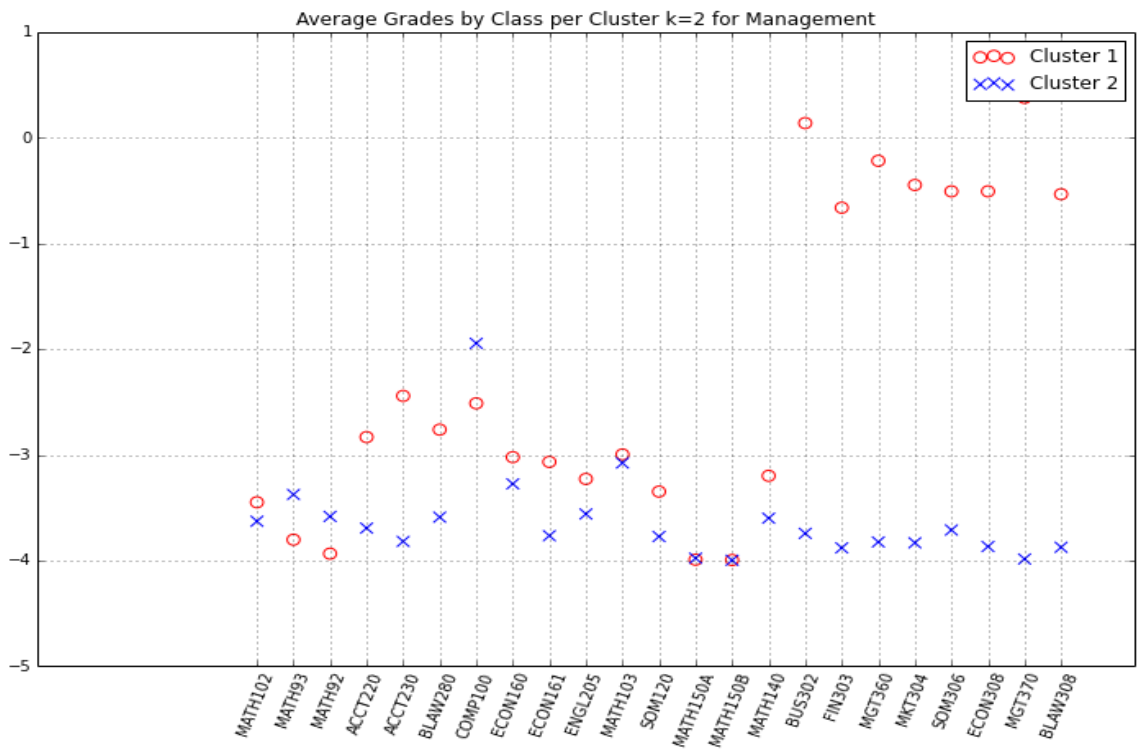